# Semiconductor Industry Trend Prediction with Event Intervention Based on LSTM Model in Sentiment-Enhanced Time Series Data

Wei-hsiang Yen[1, 2]    Lyn Chao-ling Chen[3, *]
Department of English[1]
Bachelor Program in Artificial Intelligence Applications[2]
Interdisciplinary Artificial Intelligence Center[3]
National Chengchi University

Taipei, Taiwan
lynchen@ntu.edu.tw

## Abstract

**The innovation of the study is that the deep learning method and sentiment analysis are integrated in traditional business model analysis and forecasting, and the research subject is TSMC for industry trend prediction of semiconductor industry in Taiwan. For the rapid market changes and development of wafer technologies of semiconductor industry, traditional data analysis methods not perform well in the high variety and time series data. Textual data and time series data were collected from seasonal reports of TSMC including financial information. Textual data through sentiment analysis by considering the event intervention both from internal events of the company and the external global events. Using the sentiment-enhanced time series data, the LSTM model was adopted for predicting industry trend of TSMC. The prediction results reveal significant development of wafer technology of TSMC and the potential threatens in the global market, and matches the product released news of TSMC and the international news. The contribution of the work performed accurately in industry trend prediction of the semiconductor industry by considering both the internal and external event intervention, and the prediction results provide valuable information of semiconductor industry both in research and business aspects.**

## I. Introduction

In the field of industry trend analysis, the traditional statistical analysis in business intelligence, or machine learning methods have limitations for the high variety and time series data of semiconductor industry. In the study, the research subject is the Taiwan Semiconductor Manufacturing Company (TSMC), for the leading position both in revenue and market share of semiconductor industry in Taiwan. Fluctuations in market share cause high variety of the market research documents and reports in TSMC. For example, the increasing demand of computation power from the applications of generative artificial intelligence (GenAI) facilitates the rapid innovation of the semiconductor industry, and the supply chain supply chain is reshaped for the need of specific sensors, integrated circuits and enhanced processors that influences the manufacturing decisions directly [1] [2]. Development of wafer technologies from micrometer to nanometer of sub-3 nanometer wafer processes in TSMC are considered as indicators of semiconductor industry trend in Taiwan, and the internal events of the company and the external global events were considered in sentiment analysis in the study. Public data of the quarterly financial reports on the TSMC official website can be retrieved for industry trend prediction, and the Long Short-Term Memory (LSTM) was adopted for sentiment-enhanced time series data forecasting (Figure 1). Hence, in the study, the prediction of industry trend of TSMC provides a comprehensive understanding of semiconductor industry trend in Taiwan for the great influences of TSMC on the semiconductor industry, and provides valuable information for decision making of company managers or policymaking of governments.

This paper is structured as follows: Section 2 reviews the relevant literature of the integration of artificial intelligence and the semiconductor industry, sentiment analysis and time series forecasting of financial information. Section 3 outlines the methodology that include data collection of public data on the TSMC website, training steps of sentiment analysis with event intervention and time series data forecasting, and discussions of the effect of sentiment-enhanced data in the prediction results. Section 4 concludes the study and brings topics in future work.

## II. Related works

Artificial intelligence (AI) technologies are integrated in the semiconductor industry, both facilitates manufacture including increase of production, decrease of cost, productivity improvement through precise anomaly detection and process optimization, and accelerates Time to Market (TTM) of products by AI automated tasks such as layout optimization and error detection, and improves industry trend analysis for market decision making [3] [4]. Deep learning methods provides evaluations of finance and the banking field including model preprocessing, input data, and adopted methods systematically for market decision making [5]. Machine learning methods, deep learning methods or blockchain help to improve prediction performance in business analysis and operation effectiveness [6] [7]. Graph Convolutional Network (GCN) model also performed well than the traditional models in sales prediction [8]. In credit risk prediction, FinLangNet, model uses credit loan trajectories as linguistic constructs to analyze financial event sequences, and another approach by comminating of Deep

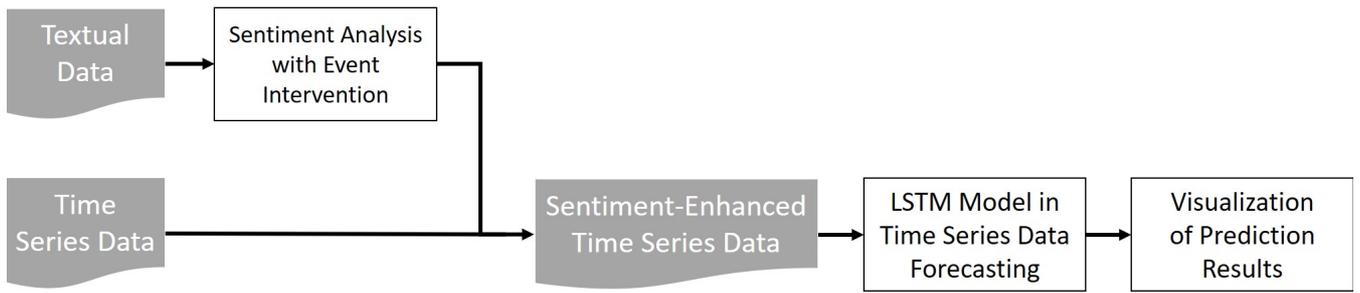

Figure 1. Data procedures in semiconductor industry trend perdition

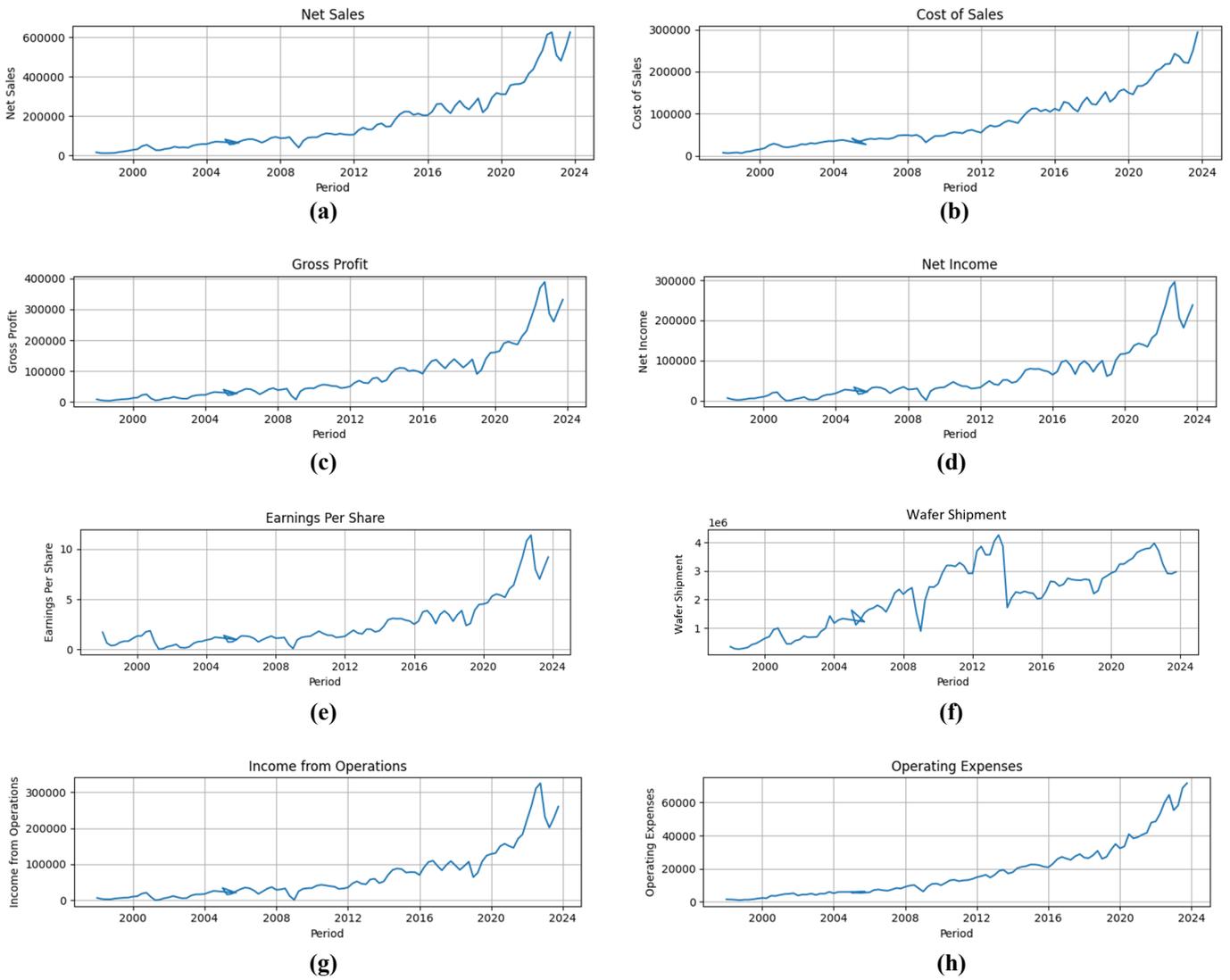

Figure 2. Time series data forecasting without event intervention from 1998 Q1 to 2023 Q4: (a) Net Sales, (b) Cost of Sales, (c) Gross Profit, (d) Net Income, (e) Earnings Per Share (EPS), (f) Wafer Shipment, (g) Income from Operations and (h) Operating Expenses.

TABLE I. POSITIVE EVENTS AND NEGATIVE EVENTS OF TSMC FROM 1998 Q1 TO 2023 Q4

| Event | Positive Event | | Negative Event | |
|---|---|---|---|---|
| | *Name* | *Weight* | *Name* | *Weight* |
| Internal | 0.25um Process<br>0.18um Process<br>0.13um Process<br>90nm Process<br>65nm Process<br>45nm Process<br>40nm Process<br>28nm Process<br>20nm Process<br>16nm Process<br>10nm Process<br>7nm Process<br>5nm Process<br>3nm Process | 1.1 | X | X |
| External | Coronavirus disease 2019 (COVID-19) | 1.2 | Internet Bubble<br>9/11 Investigation<br>SARS Outbreak<br>Global Financial Crisis (GFC)<br>Euro Area Crisis<br>Russo-Ukrainian War | 0.9 |

Recurrent Neural Network (Deep RNN) and Causal Convolution Neural Network (Casual CNN) performed well in early credit risk detection [9] [10]. However, the disadvantage of deep learning methods is the high computation cost than the traditional machine learning methods [11]. From the comparison of various deep learning methods in business process prediction, sometimes simple models outperform complex models that addresses the importance of model selection in business analytics [12].

In the aspect of time series forecasting using financial information, deep learning methods helps to retrieve features and improve accuracies of financial time series forecasting [13]. Using multiple time series data, the LSTM model and the Gated Recurrent Unit (GRU) model were adopted for financial prediction, and the temporal dependencies of the models enhance the prediction accuracy [14]. Combination the CNN model, LSTM model and the Autoregressive Moving Average (ARMA) model achieve high accuracy than the traditional models in financial time series forecasting [15]. In addition, a hybrid framework integrated wavelet transforms, stacked autoencoders and LSTM networks to capture non-linear and non-stationary features of financial data effectively in stock price prediction [16].

In the aspect of sentiment analysis using financial information, a comprehensive survey of deep learning methods in the financial field illustrates the potential and challenges in sentiment analysis [17]. CNN model performed well in sentiment analysis of financial context in stock market [18]. In addition, using zero-shot and few-shot learnings of the Large Language Model (LLM) and the BERT for Financial Text Mining (FinBERT) model, the accuracies of the sentiment classification in financial data have improved [19]. Therefore, sentiment analysis of financial reports enhances prediction of market changes to reduce risks, and improve the financial competition of market participants [20]. Hence, in the study, the LSTM model was adopted to process time series data, and sentiment analysis to process textual data for industry trend prediction of TSMC.

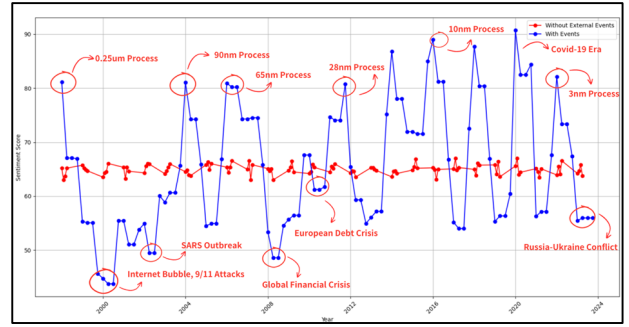

Figure 3. Event intervention enhances the results of sentiment scores, representing minor changes before event intervention, and real circumstances of TSMC after event intervention with event explanations.

III. METHODOLOGY AND RESULTS

A. Dataset

Textual data collected from the quarterly transcripts of corporate presentations on the TSMC website, and the time series data collected from the quarterly financial reports on the TSMC website (Period: 1998 Q1 to 2023 Q4, total data: 104). In the study, the LSTM model was adopted for time series data forecasting, and the data type of *Shares Outstanding* caused noises and was eliminated in the pretraining phase before event intervention from 1998 Q1 to 2023 Q4 (implemented in Python PyTorch) (Figure 2 (a) to (h)). For avoiding overfitting, 8 types of data strongly correlated to the financial and operational performance of TSMC were selected including *Net Sales*, *Cost of Sales*, *Gross Profit*, *Net Income*, *Earnings Per Share (EPS)*, *Wafer Shipment*, *Income from Operations* and *Operating Expenses* (Tabel Ⅱ). *Net sales* and *Cost of Sales* correlate to the revenue generation and production efficiency, *Gross Profit* reveals the overall financial health, *Net Income* deducts all expenses from total profit as profitability, *EPS* represents investment value, *Wafer Shipment* indicates production volume and market demand, *Income from Operations* represents core business profitability (excluding investments and taxes), and *Operating Expenses* are costs such as rent and salaries.

B. Event Intervention for Reflecting the Real World in Sentiment Analysis

Using the textual data in the collected dataset in fine-grained sentiment analysis, the results reveal the sentiment scores fluctuate between 63 and 66 (0: extremely negative to 100: extremely positive) without obvious correlation between the sentiment scores and the company states of TSMC (Figure 3). Because of the company speaker tends to declare the company state in a neutral tone for avoiding the influences on the shareholders regardless the actual company state. Therefore, in the study, the influences from internal company and external

TABLE II. SAMPLES OF SENTIMENT-ENHANCED TIME SERIES DATA WITH EVENT EXPLANATIONS FROM 1998 Q1 TO 2003 Q4 (TOTAL PERIOD: 1998 Q1 TO 2023 Q4)

| Period | Net Sales | Cost of Sales | Gross Profit | Net Income | Earnings Per Share | Wafer Shipment | Income From Operations | Operating Expenses | Sentiment Score | Event |
|---|---|---|---|---|---|---|---|---|---|---|
| 1998 Q1 | 15736.0 | 7505.0 | 8231.0 | 6947.0 | 1.7 | 350500.0 | 6709 | 1522.0 | 81.17 | 0.25um Process |
| 1998 Q2 | 11601.0 | 6304.0 | 5296.0 | 3753.0 | 0.62 | 276600.0 | 3773 | 1466.0 | 67.07 | 0.25um Process |
| 1998 Q3 | 11263.0 | 7144.0 | 4119.0 | 2115.0 | 0.35 | 263400.0 | 2827 | 1292.0 | 67.07 | 0.25um Process |
| 1998 Q4 | 11633.0 | 8010.0 | 3933.0 | 2524.0 | 0.42 | 286200.0 | 2893 | 1040.0 | 66.96 | 0.25um Process |
| 1999 Q1 | 12501.0 | 6302.0 | 6199.0 | 4090.0 | 0.68 | 319600.0 | 4873 | 1326.0 | 55.34 | 0.18um Process |
| 1999 Q2 | 17232.0 | 9696.0 | 7536.0 | 6022.0 | 0.8 | 422000.0 | 6193 | 1343.0 | 55.07 | 0.18um Process |
| 1999 Q3 | 19707.0 | 10939.0 | 8768.0 | 6137.0 | 0.81 | 465000.0 | 7169 | 1599.0 | 55.07 | 0.18um Process |
| 1999 Q4 | 23691.0 | 13797.0 | 9712.0 | 8311.0 | 1.08 | 551000.0 | 7681 | 2032.0 | 45.66 | Internet Bubble 0.18um Process |
| 2000 Q1 | 28278.0 | 15573.0 | 12705.0 | 10091.0 | 1.32 | 642000.0 | 10297 | 2408.0 | 44.75 | Internet Bubble 0.18um Process |
| 2000 Q2 | 31812.0 | 18062.0 | 13749.0 | 13349.0 | 1.33 | 697000.0 | 11490 | 2259.0 | 43.76 | Internet Bubble 0.18um Process |
| 2000 Q3 | 47491.0 | 25146.0 | 22345.0 | 20058.0 | 1.74 | 942000.0 | 18615 | 3730.0 | 43.76 | Internet Bubble 0.18um Process |
| 2000 Q4 | 53822.0 | 29066.0 | 24656.0 | 21473.0 | 1.84 | 1001000.0 | 21160 | 3596.0 | 55.43 | Internet Bubble 9/11 Investigation 0.13um Process |
| 2001 Q1 | 39521.0 | 26043.0 | 11089.0 | 8420.0 | 0.71 | 702000.0 | 9257 | 4221.0 | 55.43 | Internet Bubble 9/11 Investigation 0.13um Process |
| 2001 Q2 | 26298.0 | 21299.0 | 4999.0 | 312.0 | 0.01 | 450000.0 | 284 | 4714.0 | 51.05 | Internet Bubble 9/11 Investigation 0.13um Process |
| 2001 Q3 | 26940.0 | 20124.0 | 6816.0 | 1237.0 | 0.06 | 448000.0 | 1942 | 4874.0 | 51.05 | Internet Bubble 9/11 Investigation 0.13um Process |
| 2001 Q4 | 33130.0 | 22041.0 | 11089.0 | 4514.0 | 0.26 | 558000.0 | 5859 | 5230.0 | 53.83 | No significant event |
| 2002 Q1 | 35790.0 | 23763.0 | 12027.0 | 6588.0 | 0.35 | 599000.0 | 8182 | 3845.0 | 54.92 | No significant event |
| 2002 Q2 | 44182.0 | 27759.0 | 16423.0 | 9310.0 | 0.49 | 71900.0 | 11980 | 4448.0 | 49.49 | No significant event |
| 2002 Q3 | 39835.0 | 27000.0 | 12835.0 | 3160.0 | 0.16 | 677000.0 | 8300 | 4470.0 | 49.49 | No significant event |
| 2002 Q4 | 41154.0 | 30272.0 | 10682.0 | 2553.0 | 0.13 | 682000.0 | 5651 | 5031.0 | 60.06 | SARS Outbreak 90nm Process |
| 2003 Q1 | 39325.0 | 28939.0 | 10368.0 | 4385.0 | 0.23 | 690000.0 | 6195 | 4191.0 | 58.87 | SARS Outbreak 90nm Process |
| 2003 Q2 | 49922.0 | 31571.0 | 18351.0 | 11730.0 | 0.58 | 887000.0 | 13340 | 5011.0 | 60.64 | SARS Outbreak 90nm Process |
| 2003 Q3 | 54877.0 | 33430.0 | 21447.0 | 15169.0 | 0.75 | 992000.0 | 16487 | 4960.0 | 60.64 | SARS Outbreak 90nm Process |
| 2003 Q4 | 57780.0 | 35072.0 | 22707.0 | 16002.0 | 0.79 | 1427000.0 | 16625 | 6082.0 | 65.69 | 90nm Process |

global events on TSMC were considered for revealing the circumstances of TSMC in the real world.

Events intervention has defined according to the news between 1998 to 2023 (Table I), internal events including generations of chip manufacturing of TSMC were considered as positive events including the wafer process of TSMC from 0.25um process in 1998 to the 3nm process in 2022; otherwise, events may harm for TSMC were considered as external negative events including internet bubble, 9/11 investigation, Severe Acute Respiratory Syndrome (SARS) outbreak, Global Financial Crisis (GFC), Euro area crisis and Russo-Ukrainian War. In addition, Coronavirus disease 2019 (COVID-19) was considered as an external positive event of TSMC for the increasing demand of chips. Weights of the events were set according to the positive influence or negative influence and degrees of influences of the events on TSMC (Table 1), for example the COVID-19 event (Weight: 1.2) has grater positive influences than the other positive events (Weight: 1.1). After event intervention, the positive events (Weight: 1.1) increase the original sentiment scores, and negative events (Wight: 0.9) decrease the original sentiment scores (Figure 3). From the results of sentiment analysis with event intervention (Figure 3), there are high peaks matching the positive events of wafer processes of TSMC and the COVID-19 event, and there are troughs matching all negative events. Hence, event intervention enhances the results of semimetal analysis, and reveals the real circumstances of TSMC by both considering the positive or negative internal influences of the events in the company and the external influences of the global events (implanted in Python TextBlob).

*C. LSTM Model for Sentiment-Enhanced Time Series Data Forecasting*

Sentiment-enhanced time series data includes the 8 types of time series data (*Net Sales*, *Cost of Sales*, *Gross Profit*, *Net Income*, *Earnings Per Share (EPS)*, *Wafer Shipment*, *Income from Operations* and *Operating Expenses*), and *Sentiment*

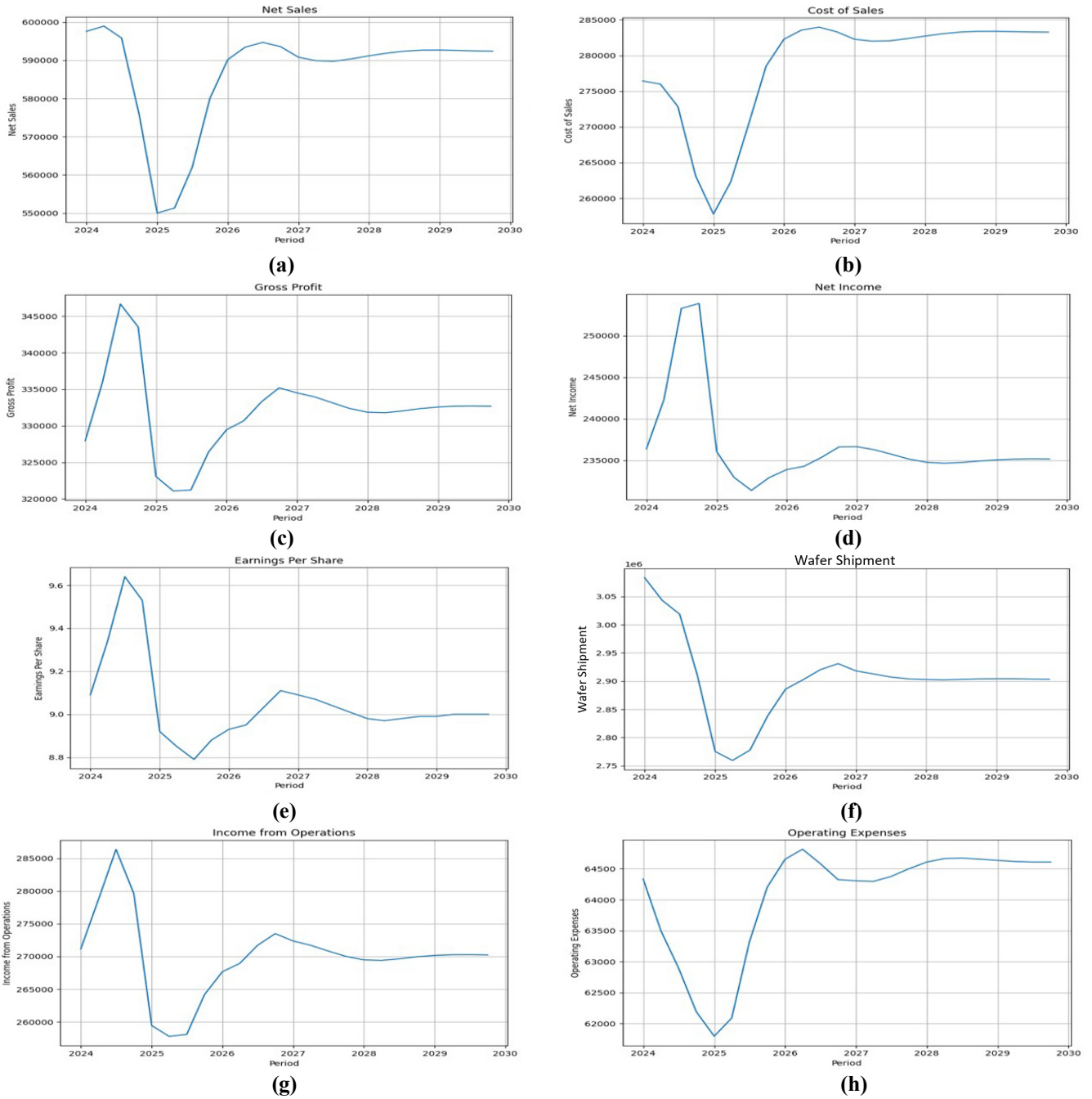

Figure 4. Time series data forecasting with event intervention from 2024 Q1 to 2029 Q4: (a) Net Sales, (b) Cost of Sales, (c) Gross Profit, (d) Net Income, (e) Earnings Per Share (EPS), (f) Wafer Shipment, (g) Income from Operations and (h) Operating Expenses.

*Scores* from the sentiment analysis were also considered (Tabel Ⅱ), and the LSTM model was adopted to predict the period from 2024 Q1 to 2029 Q4 (implanted in Python PyTorch). From the prediction separately, curves of the 8 types of data drop significantly in 2025, rebound in 2026 to 2027, and there are flat periods results of 8 types of sentiment-enhanced time series data between 2027 to 2029 (Figure 4 (a) to (h)). For the event intervention enhance the sentiment analysis, and also reveals in the sentiment-enhanced time series data. From the prediction results of combinations of 8 types of sentiment-enhanced time series data, the peak appears in 2024 Q4 (Figure 5), and matches the 2-nanometer released news of TSMC that the 2-nanometer process is in the trial production phase in 2024, and plans to mass production in 2025 [21]. The peak appears in 2027 Q1 (Figure 5), and matches the 1-nanomete product released news of TSMC that the 1-nanometer process will enter production in 2027 in a leading position of advanced semiconductor technology [22]. The trough appears in 2025 Q3, and may cause by the yield problems of production issues or breakthroughs from competitors [23]. In addition, the policy of previous United

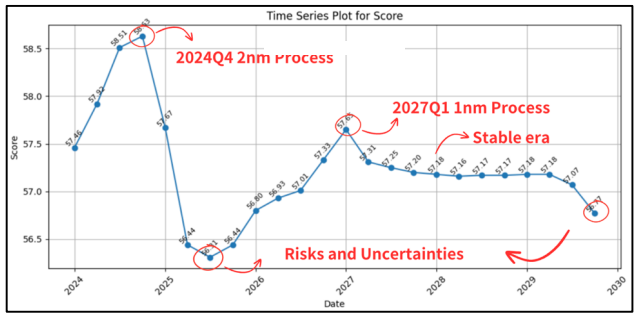

Figure 5. Time series data forecasting after event intervention from 2024 Q1 to 2029 Q4 with event explanations, revealing peaks matching the product released news of TSMC, and the trough matching the international news of the potential threats.

States (US) government increases US reliance on chips of TSMC, however, the new policy from current US government may restrict the expansion of TSMC in US [24]. Hence, using sentiment-enhanced time series data, the prediction results of LSTM model performed well, and the correlated news provides the evidences.

## IV. CONCLUSION AND FUTURE WORKS

An approach proposed by considering the event intervention in sentiment analysis, and the sentiment-enhanced time series data performed well in industry trend forecasting in the semiconductor industry. In the future work, the approach can be applied in various business analysis for providing valuable information in short-term or long-term decision making. In the future works, more business cases in various industries can be evaluated for comparison and discussion.

## ACKNOWLEDGMENTS

This work was partially supported by the National Science and Technology Council, Taiwan, under grants 114-2635-E-004 -001- and 113-2813-C-004-031-E. Special thanks for the reporter, Yi-cheng Lai.